\documentclass[runningheads]{llncs}
\usepackage{graphicx}
\usepackage{hyphenat}
\usepackage{tabularx}
\usepackage{comment}
\usepackage{framed}
\usepackage{tabularx}
\usepackage{graphicx}
\usepackage{amssymb}
\usepackage{tabularx}
\usepackage{threeparttable}

\newsavebox{\twosubbox}
\usepackage{dirtytalk}
\usepackage{csquotes}
\usepackage{tabulary}
\usepackage{float}
\usepackage{subcaption}
\usepackage{ragged2e}
\usepackage{url}

\begin{document}

\title{Curatr: A Platform for Semantic Analysis and Curation of Historical Literary Texts}

\titlerunning{Curatr: A Platform for Curation of Historical Texts}

\authorrunning{Leavy et al.}

\author{Susan Leavy  \and Gerardine Meaney \and
Karen Wade \and Derek Greene}

\institute{University College Dublin, Ireland}


\author{Susan Leavy \and Gerardine Meaney \and Karen Wade \and Derek Greene}

\authorrunning{S. Leavy et al.}
%
\institute{University College Dublin, Ireland\\
\email{\{susan.leavy,gerardine.meaney,karen.wade,derek.greene\}@ucd.ie}}

\maketitle

\begin{abstract}

The increasing availability of digital collections of historical and contemporary literature presents a wealth of possibilities for new research in the humanities. The scale and diversity of such collections however, presents particular challenges in identifying and extracting relevant content. This paper presents \emph{Curatr}, an online platform for the exploration and curation of literature with machine learning-supported semantic search, designed within the context of digital humanities scholarship. The platform provides a text mining workflow that combines neural word embeddings with expert domain knowledge to enable the generation of thematic lexicons, allowing researches to curate relevant sub-corpora from a large corpus of 18th and 19th century digitised texts.




\end{abstract}

\section{Introduction}

The interpretability of the algorithmic process and the incorporation of domain knowledge are essential to the use of machine learning and text mining in the semantic analysis of literature. The absence of these factors can inhibit adoption of machine learning approaches to text mining in the humanities, due to issues of accuracy and trust in what is often regarded as a `black-box' process~\cite{chiticariu2013rule,frank2012semantic,hampson2013improving}. This paper presents \textit{Curatr}, an online platform that incorporates domain expertise and imparts transparency in the use of machine learning for literary analysis. The system supports a corpus curation workflow that addresses the requirements of scholars in the humanities who are increasingly working with large collections of unstructured text and facilitates the development of sub-corpora from large digital collections. 

Selection, curation and interpretation is central to knowledge generation in the humanities~\cite{jockers,wolfe2008annotations}. This process is supported in \textit{Curatr} with conceptual search functionality that uses neural word embeddings to build conceptual lexicons specific to a given theme or topic. These thematic lexicons can then be used to mine relevant texts to form curated literary collections that may be saved, further modified, or exported as sub-corpora. The platform was developed based on a collection of 35,918 English language digital texts from the British Library\footnote{British Library Labs: \url{https://www.bl.uk/projects/british-library-labs}}. 
 
 An evaluation of \emph{Curatr} was conducted in conjunction with an associated project examining the relationship between societal views of migration, ethnicity, and concepts concerning contagion and disease in 19th century Britain and Ireland. In order to explore the cultural representation of migrants, the study focused on their representation within historical fiction which comprises 16,426 texts of the British Library digital collection. Given the largest communities of migrants to London during the late 19th century were Irish and Jewish, this study focused on their portrayal in relation to prevailing concepts of contagion, disease and migration. Lexicons related to these themes were generated through recommendations derived from word embedding models and text were retrieved based on relevance of the texts. The findings were evaluated in terms of the overall requirements of a humanities scholar along with the relevance of the texts uncovered. We describe this case study in more detail in Section \ref{sec:case}.




\section{Related Work}

\subsection{Text Mining in the Humanities}



The work builds on a range of literature that demonstrates requirements for digital humanities platforms.  Close reading functionality to provide context is an essential aspect of humanities research, as evidenced in the provision of close reading functionality along with quantitative analysis in systems developed by Hinricks et al.~\cite{hinrichs2015trading} and Vane~\cite{vane2018text}. Domain knowledge was combined with automated text classification to provide more accurate retrieval results in a system developed by Sweetnam and Fennel~\cite{bailey2012cultura} to explore early-modern English texts. A system based on semantic search was demonstrated by Kopaczyk et al~\cite{kopaczyk2013legal} for the analysis of Scottish legal documents from the 16th century. Other systems, such as that proposed by Jockers~\cite{jockers}, have focused primarily on the use of machine learning methods for text analysis. 

In a study of the update of machine learning in industry, Chiticariu et al.~\cite{chiticariu2013rule} noted a gap in the volume of academic research on machine learning, compared with lower levels of uptake within industry and found the causes of this pertained to training data, interpretability and incorporation of domain knowledge. Similarly, the relatively low uptake of machine learning methods in the digital humanities has been attributed to issues pertaining to interpretation and trust~\cite{vanCranenburgh2019,frank2012semantic,hampson2013improving}. Imparting domain knowledge into the process of text analysis through interpretation and annotation is also central to humanities research ~\cite{jackson2002marginalia,wolfe2008annotations}. The \emph{Curatr} platform addresses these specific requirements of digital humanities research by incorporating domain knowledge and transparency within a text mining workflow.


\subsection{Concept Modelling with Word Embeddings}

\emph{Word embedding} refers to a family of methods from natural language processing that involve mapping words or phrases appearing in large text corpora to dense, low-dimensional numeric representations. Typically, each unique word in the corpus vocabulary will be represented by its own vector. By transforming textual data in this way, we can use the new representation to capture the semantic similarity between pairs or groups of words. Word embedding methods have been used in digital humanities research to generate semantic lexicons for a range of purposes including detecting language change over time ~\cite{hamilton2016inducing}, extracting social networks from literary texts~\cite{wohlgenannt2016extracting}, sentiment analysis ~\cite{tang2014learning}, and semantic annotation~\cite{leavyindustrial}. An interactive strategy whereby a user incrementally creates a lexicon based on recommendations for similar words as recommended by a word embedding model has been demonstrated in a number of works~\cite{fast2016empath,park2018conceptvector}.

A variety of different approaches have been proposed in the literature to construct embeddings. The word embedding algorithm used in this research is \emph{word2Vec}~\cite{mikolov13efficient}, which generates distributed representations of words that can be used to interpret their meaning. This approach captures the concept from distributed semantics that the meaning of a word ``can be determined by the company it keeps''~\cite{firth57synopsis}. Word co-occurrence is identified over an entire corpus and each word along with the words found beside it in the text are represented by a vector. The similarity of terms can then be derived based on whether they are used alongside similar words, or in a similar context. This approach to generating lexicons has been shown to be useful where the language of a particular corpus is highly specific~\cite{chanen2016deep} and where existing general-purpose lexicons are not appropriate and is therefore particularly relevant to digital humanities. It has also been pointed out that pre-processing and methods of representing text can have particular significance within a digital humanities context~\cite{camacho2017role,flanders2018shape}. These decisions are a crucial aspect of the evaluation of results within humanities scholarship. Given that using word embedding in text analysis has been critiqued for its lack of transparency~\cite{subramanian2018spine}, it is crucial within a digital humanities context to impart transparency into the text mining workflow.

\section{Curatr Design}


\subsection{Platform Overview}
\begin{figure}[!t]
 \centering
\includegraphics[width=0.95\linewidth]{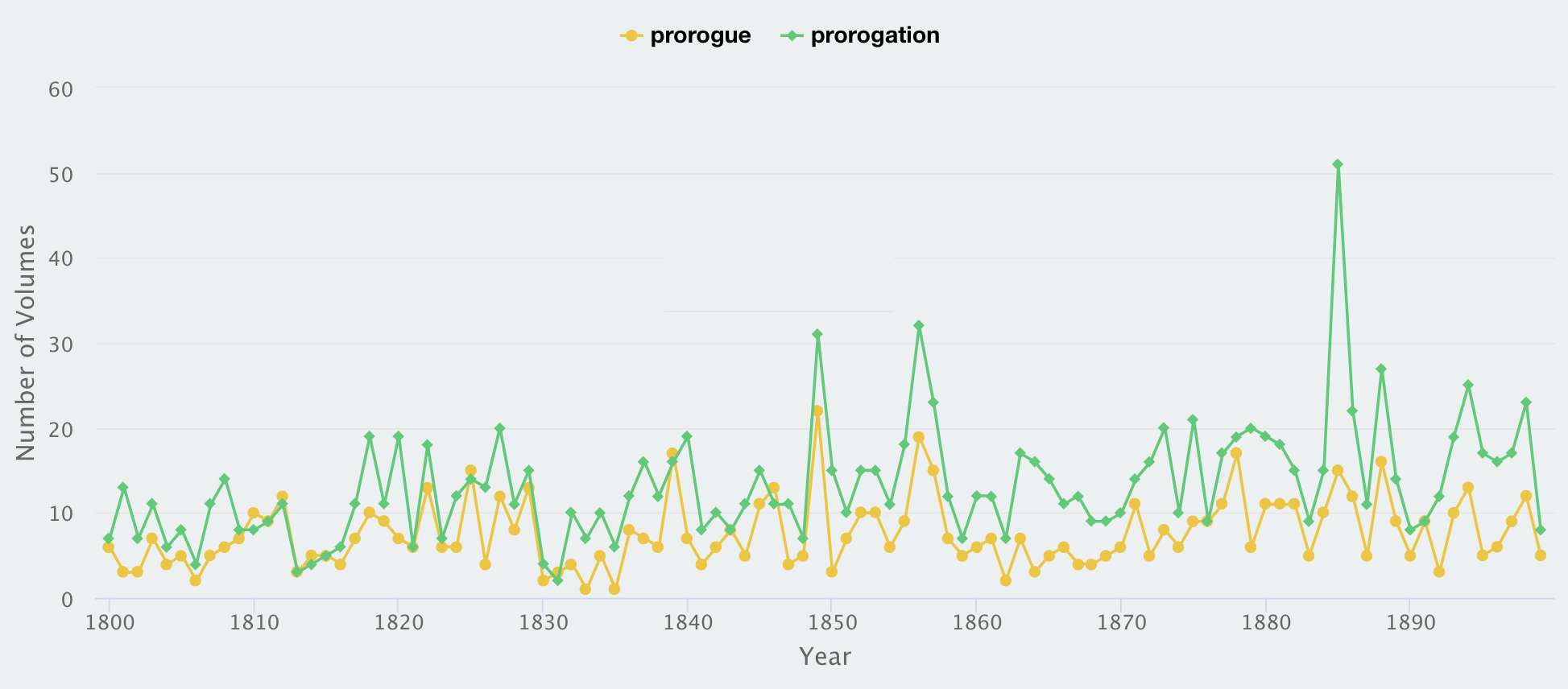}
  \caption{Sample \textit{Curatr} n-gram frequency analysis.}
  \label{ngrams}
\end{figure}

\textit{Curatr} implements a text mining framework involving conceptual search using word embeddings to dynamically build a semantic lexicon specific to a given literary corpus.  The humanities researcher begins with seed terms and these are expanded using neural word embedding through an interactive online interface to produce semantic lexicons. The \emph{Curatr} system also provides for keyword search, filtering based on metadata, ngram frequencies, and categorisation based on the original British Library topical classifications (Fig. \ref{ngrams}). The information retrieval component of the system involves the indexing of texts and using the open source Apache Solr engine\footnote{\url{http://lucene.apache.org/solr/}}.


\begin{figure}[t]
 \centering
  \includegraphics[width=.75\linewidth]{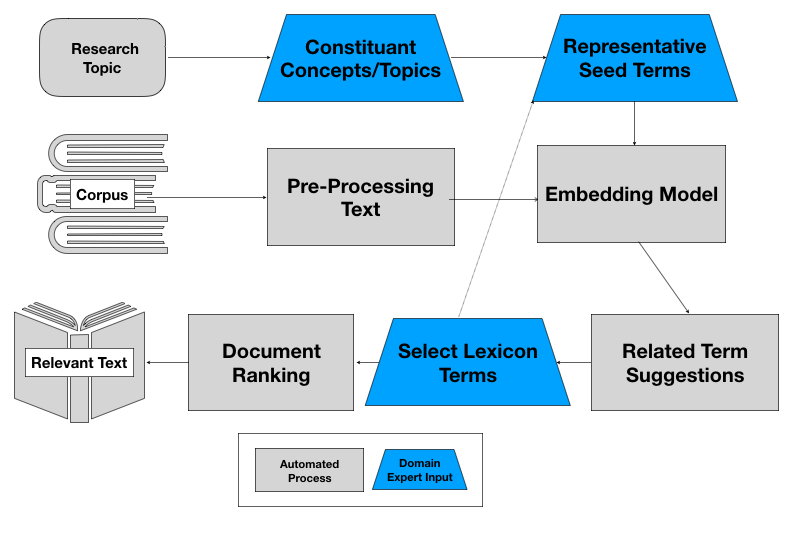}
  \caption{\emph{Curatr} workflow for digital humanities text mining.}
  \label{workflow}
\end{figure}



\subsection{Concept Lexicon Generation}

The conceptual search workflow outlined in Fig. \ref{workflow} enables the compilation of seed terms associated with a given concept or topic by the humanities researcher. \emph{Curatr} allows for the expansion of these terms to form a lexicon of semantically similar words based on associations suggested from the querying of a neural word embedding model developed from the corpus.  Word embedding models were generated from the complete English language corpus using the \emph{word2vec} approach~\cite{mikolov13efficient} yielding real-valued, low-dimensional representations of words based on lexical co-occurrences. The use of word embeddings rather than more complex language models were deemed appropriate due to the lack of structure in the text and OCR errors introduced through the process of digitisation. The specific embedding variant used in this work is a 100-dimensional Continuous Bag-Of-Words (CBOW) \emph{word2vec} model, trained on the full-text volumes of the corpus. To address the levels of transparency required in digital humanities scholarship regarding approaches to text representation and parameter settings and their potential effects on results, decisions regarding text processing options and parameter strategies in generating the neural word embedding models are available to the user.

Based on the embedding model, the top 20 words found to be similar to the seed words are recommended to expand the current lexicon. The user selects the subset of recommendations to be included in the conceptual lexicon. This inclusion of a ``human in the loop'' ensures that the process of generating a lexicon is informed by domain knowledge of the user. Multiple iterations of this semantic search allow the researcher to refine the lexicon and augment the conceptual category on each iteration, simulating the process of knowledge generation from close reading and annotation.


\subsection{Corpus Curation}

The finalised semantic lexicons are used as a basis for ranked volume retrieval from the indexed library corpus. Texts are ranked according to the frequency of occurrence of terms from each generated lexicon relative to the document length. This process uncovers documents that pertain to the given conceptual category. These document rankings may then be saved by the user for later use.



\textit{Curatr} also facilitates the amendment of the sub-corpus to remove non-relevant texts. Based on the results of this search, the user may revise the contents of the conceptual lexicon and rerun the text mining process. Along with the provision of a close reading interface for examining individual texts, the platform also enables the export of curated sub-corpora to download for further analysis.


\section{Case Study Evaluation}
\label{sec:case}

The case study involved an exploratory investigation of historical cultural attitudes towards migration in Britain and their associations with concepts of contagion and disease from the set of 16,426 texts in the corpus ~\cite{nelkin1988placing,kinealy2006great}. Poverty-induced migration from Ireland to Britain during the Great Famine (1845-1851) is cited as generating a fear of transmission of contagious disease~\cite{morash2009hungry}. The association of fear of contagion with historical sentiments towards migrants, according to Samuel K. Cohn however, requires more systematic study and more nuanced interpretation of both historical and cultural archives~\cite{cohn2012pandemics}.  The aim of this exploratory study is to uncover new texts and literary representations that are less well known to researchers and which might challenge current understandings of the relationship between migration and concepts of contagion and disease.

Evaluating the usefulness of retrieved documents in the context of humanities research does not always align with standard information retrieval precision metrics. This is particularly pertinent in the preliminary exploratory phase of humanities research, where new and diverse texts are sought by the researcher and their relevance to the research topic is often not immediately apparent~\cite{clarke2008novelty}. However, a prime advantage of text mining in the humanities is the potential to uncover novel and diverse texts that could challenge prevailing theory. This research therefore draws on Bates et al.~\cite{bates1996getty}, who examined what constitutes relevance of a document in the context of humanities research defining two aspects of relevance in humanities research, \emph{content relevance} and  \emph{utility relevance}. Content relevance determines whether the search terms in the query relate to the content retrieved while utility relevance evaluates the usefulness in the context of the research topic. Given that a humanities researcher requires texts that inform their topic, texts deemed to be most relevant in terms of utility are often not those with the highest content relevance, but are those that alter the judgement and challenge existing theories of the researcher. Document unfamiliarity is therefore an attribute that `may be presumed to swamp all other considerations'~\cite[p.~702]{bates1996getty}.  Unfamiliarity of a text was defined by Barry et al. (~\cite{barry1994user})  in terms of content, source or stimulus. \emph{Content novelty} describes how the content of a text itself provides new knowledge to a researcher. \emph{Source novelty} refers to whether the text is written by a previously unknown author or publisher.  \emph{Stimulus novelty} indicates whether the text itself is new to the researcher. Brought together the framework outlines capture the complexity of judgements of value and relevance of text mining for the humanities.

\begin{figure}[!t]
 \centering
    \fbox{\includegraphics[width=.75\linewidth]{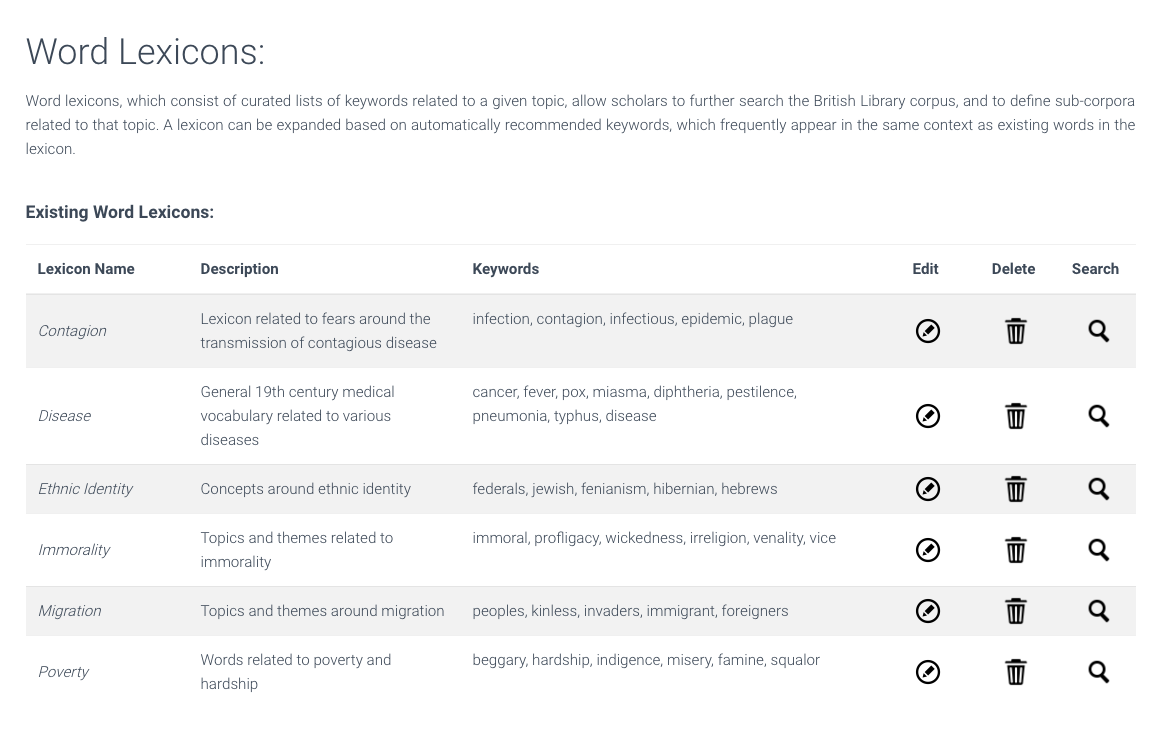}}
  \caption{The \emph{Curatr} lexicon management interface.}
  \label{semanticsearch}
\end{figure}

The key thematic strands of the case study were identified and associated seed terms defined.  The themes pertained to `migration', `illness', `contagion', and Irish and Jewish ethnic identities. The word embedding model was queried to uncover words used in a similar context to the seed terms in the fiction corpus (see Fig. \ref{semanticsearch}). The results were selectively added based on contextual knowledge and interpretation on the part of the humanities researcher through an interactive lexicon building interface. The resulting expanded semantic lexicons were then used as a basis for ranking the documents to uncover texts that capture the key themes of the study. See Table \ref{expandedset} for examples of lexicons generated in this way.

\begin{table*}[!b]
\scriptsize
\begin{tabularx}{\linewidth}{>{\hsize=.18\hsize\RaggedRight}X>{\hsize=.27\hsize\RaggedRight}X>{\hsize=.62\hsize}X}

   
    \textbf{Lexicon} & \textbf{Seed Terms} & \textbf{Recommended Words} \\ \hline
  
    \emph{Ethnic identity} & irish, fenian, papist, jewish, jew & jews, fenianism, hibernian, usurer, celtic, rebels, invincibles, whiteboys, incendiary, brogue, chartists, irishman, catholics, ringleaders, rabbis\\
    
    \emph{Migration} &  immigrant, alien, interloper, migrant   & civilisers, doavn trodden, self governing, circumcised, peoples, separatists, cousinhood, usurpers, interloper, aliens, intruder, middlemen, ryots, kinless, alien\\
    \emph{Contagion} & infect, epidemic, inoculate, contagion, contaminate, vaccinate & infection, contagion, infectious, infected, fever, contagious, epidemic, plague, epidemic, disease, epidemics, fever,  malarial, endemic,  malady\\
    \emph{Disease} & disease, smallpox, cholera, fever, pestilence & scarlet fever, epidemics, morbus, cancers, tumour, incurable, malaria, sickness, cancer, brain, diseases, distempers, typhus, anaemia, heart disease\\
  \hline
  \end{tabularx}
  \vskip 5px
   \caption{Examples of top recommended words for different conceptual lexicons.}
  \label{expandedset}
\end{table*}



Analysis of the recommended terms suggested new associations in the fiction corpus. For example, terms pertaining to the theme of civil unrest \textit{(rebels, dynamiters, conspirators, incendiarism, anarchistic, nihilists, revolutionary, sedition, informers, radicals}) make up a substantial portion of the lexicon pertaining to `ethnic identity'. As a result of these suggestions, aligning with the iterative theory-building approach of research in the humanities, the suggested words enriched the original conceptualisation of the research theme of ethnic identity to incorporate relationships with political ideology. Additionally, the expanded lexicons indicated points where conceptual ontologies overlap. Suggested terms in \emph{Curatr} also provided indications of relevant but unfamiliar or archaic terminology. For instance, in this study the term `distemper', which is more commonly applied to animal illnesses nowadays, appears as a term that is relevant to the concept of contagion within these works. OCR (optical character recognition) and spelling errors and variants were also uncovered using this method. 

\subsection{Finding and Discussion}

Drawing upon the framework of retrieval value outlined by Bates et al.~\cite{bates1996getty} and Barry et al.~\cite{barry1994user}, the top 10 retrieved texts in each category were analysed in terms of novelty to the humanities researcher involved in the project. The qualitative difference in text retrieved before and after the query expansion phase was also examined. Texts in the corpus were ranked according to the frequency by which the curated conceptual lexicons were mentioned, relative to the length of the documents.  The iterative process whereby the researcher returns to edit the conceptualisation of the key thematic trends of the research topic through editing the semantic lexicons aligns with the process of grounded theory research and necessitated a close reading of the results by humanities researchers to evaluate the value of the retrieved documents.

\begin{table*}[!t]
\scriptsize
\begin{tabularx}{\linewidth}{>{\hsize=.6\hsize\RaggedRight}X>{\hsize=.6\hsize\RaggedRight}X}

\multicolumn{1}{c}{\textbf{Prior to Query Expansion}} & \multicolumn{1}{c}{\textbf{Post Query Expansion}}\\
\hline

1894 The Captain's Youngest,  Frances H.  Burnett\newline
1888 The Devil`s Die,  Allen Grant\newline
1894  The Azrael of Anarchy,  Gustave Linbach\newline
1891  The Year of Miracle,  Fergus	Hume\newline 
1870  Unawares,  Frances Mary 	 Peard\newline
1893 Doctor, or Lover?, Faber Vance\newline
1888 A Crown of Shame, Florence Marryat\newline
1875 Ashes to ashes,  Hugh R. Haweis\newline
1865 Not Proven, Christina B Cameron\newline
1892 The Medicine Lady,  Elizabeth T Meade\newline
&
1891 The Year of Miracle, Fergus Hume\newline
1897 The Sign of the Red Cross, Evelyn E. Green\newline
1855 Old Saint Paul`s, William H. Ainsworth\newline
1894 The Azrael of Anarchy, Gustave Linbach \newline
1847 A Tale of the Irish Famine, Unknown\newline
1885 The Legend of Samandal, James Fer\newline
1855 The Wood-Spirit, Ernest C. Jones\newline
1888 The Devil`s Die, Grant Allen\newline
1898 Vanya, Orlova Olga \newline
1897 A Literary Gent, J.C. Kernahan\newline \\
\end{tabularx}
\caption{Texts retrieved before and after query expansion, for concepts related to `illness' and `contagion'.}
\label{illnessresults}
\end{table*}

\begin{table*}[h]
\scriptsize
\begin{tabularx}{\linewidth}{>{\hsize=.6\hsize\RaggedRight}X>{\hsize=.6\hsize\RaggedRight}X}
\multicolumn{1}{c}{\textbf{Prior to Query Expansion}} & \multicolumn{1}{c}{\textbf{Post Query Expansion}}\\
\hline
1898 For Lilias, Rosa Nouchette Carey\newline 
1875 The Golden Shaft, George C. Davies\newline 
1876 The Youth of the Period,  James F.S. Kennedy\newline 
1876 Her Dearest Foe, Alexander\newline 
1886 A Modern Telemachus, Charlotte M. Yonge\newline 
1887 Major Lawrence, Emily Lawless\newline 
1857 Guy Fawkes, William H. Ainsworth\newline 
1846 The Moor, the Mine and the Forest, William Heatherbred\newline 
1865 The Notting Hill Mystery, Charles Felix\newline 
1897 Owen Tanat, Alfred N. Palmer\newline 
&
1881 Gifts and Favours Doctor Olloed, Unknown\newline 
1871 Ierne, William R. Trench\newline 
1863 Sackville Chase,  Charles J. Collins\newline 
1894 Ivanda or the Pilgrim's quest, Claude A. Bray\newline 
1894 Doctor Izard, Anna K. Green\newline 
1865 The Crusader or the Witch of Finchley, Unknown\newline 
1864 The Bee-Hunters, Gustave Aimard\newline 
1850 Helen Porter, Thomas P. Prest\newline 
1852 Idone, James H. L. Archer\newline 
1886 A Mysterious Trust, Edmund Mitchell\newline 
\\
\end{tabularx}
\caption{Texts retrieved before and after query expansion, for a concept related to `migration'.}
\label{migrationresults}
\end{table*}

\subsubsection{Source novelty.}

A striking pattern among the texts retrieved with the expanded query set was the relative obscurity of some of the authors. In analysing the top ten texts retrieved using the seed terms alone, five of the migration category and four of illness category were immediately familiar to the researchers (Tables \ref{illnessresults} and \ref{migrationresults}). However, after the list was expanded based on suggestions from neural word embedding, this figure reduced to three for the concept of illness, and none were previously well known in relation to the concept of migration. 

A similar trend is evident in the texts retrieved using the semantic lexicons of ethnic identity where only one of the authors in the top ten texts related to Irish identity and three relating to Jewish identity were widely known (see Table. \ref{retrivedtext_ethnic}).  The texts retrieved based on the semantic lexicon of Irish ethnicity contained none by widely read Irish authors, although Ulick Ralph Burke and Albert Stratford George Canning do merit an entry in the database of Irish Literature\footnote{Ricorso: Database of Irish writers \url{http://www.ricorso.net}}. A work by the prolific English author Fergus Hume was retrieved, as were works by the British authors Mabel Collins and Edith Cuthell, the latter novel being its author's only work based in an Irish setting. This identification of a number of writers less prominent in the canon, but who wrote on topics relating to the key themes of the case study, demonstrate how the expansion of query terms in this context has the potential to uncover relevant texts which might otherwise be overlooked.

\subsubsection{Stimulus novelty.}

\begin{table*}[h]
\scriptsize
\centering
\begin{tabularx}{\linewidth}{>{\hsize=.5\hsize\RaggedRight}X>{\hsize=.5\hsize\RaggedRight}X}
\multicolumn{1}{c}{\textbf{Irish Lexicon}} & \multicolumn{1}{c}{\textbf{Jewish Lexicon}}\\
\hline
1897 Sweet Irish Eyes, E. Cuthell\newline 
1893 The Harlequin Opal, F. Hume\newline 
1875 The Autobiography of a Man-o-War`s Bell, C. R. Low\newline 
1883 The Wild Rose of Lough Gill, P. G. Smyth\newline
1880 Loyal and Lawless, U. R. Burke\newline 
1867 Baldearg O'Donnell, A. A. G. Canning\newline 
1893 The Great War of 189-. A forecast, F. Villiers\newline 
1896 The Idyll of the White Lotus,  M. Collins\newline 
1886 Our Radicals. A tale of love and politics, F. Burnaby\newline 
1891 The Last Great Naval War, A. N. Seaforth\newline 
1890 Heir and no Heir Canning,  A. S. G. Canning\newline 
&
1890 The Prophet. A parable, T. H. H. Caine\newline 
1897 A Rogue`s Conscience, D. C. Murray\newline 
1864 The Hekim Bashi, H. Sandwith\newline 
1874 Jessie Trim, B. L. Farjeon\newline 
1853 The Turk and the Hebrew, Unknown\newline 
1865 The Crusader or the Witch of Finchley, Unknown\newline 
1895 Maid Marian and Crotchet Castle, T. L. Peacock\newline 
1898 Dreamers of the Ghetto, I. Zangwill\newline 
1838 Oliver Twist, C. Dickens\newline 
1869 Count Teleki: A story of modern Jewish life, UN\newline 
\end{tabularx}
\caption{Texts retrieved for the `Irish' and `Jewish' conceptual lexicons.}
\label{retrivedtext_ethnic}
\end{table*}

In this case study the texts retrieved presented a number of works that represented not only less well-known authors of the 18th and 19th centuries, but also less well-known texts written by more prominent authors. In relation to the texts retrieved using the conceptual lexicon of illness for instance, the majority of retrieved titles were not commonly known and none pertaining to Irish identity attracted significant bibliographic or critical commentary. The text marked as most relevant to Irish identity, \emph{Sweet Irish Eyes}, was written by an English author and was their only Irish-themed work. Thee expanded list of query terms also uncovered slightly older texts pointing to the potential of word embedding to compile a semantic lexicons that are less biased towards contemporary linguistic norms. 

The most relevant of the retrieved set of texts pertaining to Jewish identity was that by British novelist Hall Caine. This work is referred to in studies of the Gothic (e.g. Mulvany-Roberts~\cite{mulvey2016handbook}), but there is very little reference to this novel and its portrayal of Jewish characters. It is notable that Anglo-Jewish writers (Zangwill and Farjeon\footnote{Farjeon features more prominently in histories of Australian literature as he emigrated there.}) feature prominently alongside one of Dickens' best known novels. Charles Dickens' \emph{Oliver Twist} and Israel Zangwil's \emph{Dreamers of the Ghetto}, which appear on the list of novels pertaining to Jewishness, were the only two novels on the list that have previously attracted sustained scholarly attention. For example, Udelson~\cite{udelson1990dreamer}, Rochelson~\cite{rochelson1999they,rochelson2010jew} and Murray~\cite{murray2015social} have all written on Zangwill, though he is far less well known to the public than Dickens. 




\subsubsection{Content novelty.}

While uncovering authors and titles that were previously less familiar to a researcher is beneficial, its value is ultimately dependent on the relevance of the content. Close analysis showed that the expanded set of query terms retrieved books that were less specific to the topics than those retrieved using the manually generated set of seed terms suggesting new associations and challenging the researcher's understanding of the key themes. For instance, in relation to the concept of illness, a text pertaining to the Irish Famine and texts addressing political issues were returned. The most relevant to the concept of illness using an expanded lexicon was Evelyn Everett Green's novel, \emph{The Sign of the Red Cross}. The content of this novel describes details of a plague in London and was critiqued at the time for its graphic nature~\cite{dempster1983thomas}. The novel \emph{Old Saint Paul's} was concerned with the plague in London and provides a wealth of information relevant to the topic. While these novels are not among the most well known currently, their graphic account of plague in London provides a wealth of insight regarding the historical conceptualisation of illness in Britain. The content of texts retrieved pertaining to Irish and Jewish ethnic identities were highly relevant to the topic and uncovered texts that are less well known, thus presenting new opportunities to examine cultural representations of ethnicity. For instance, the theme of seafaring in the titles pertaining to Irish identity is an association that was new to the researchers.  The content that was returned in relation to the concept of migration returned more titles that addressed migration in the broader sense in connection with international political events, in contrast with the narrower focus of the results retrieved using seed terms alone.

\section{Conclusion}

This research presents an approach whereby machine learning and text analysis is used within a text mining workflow designed for digital humanities research. The corpus curation workflow in \emph{Curatr} is supported by neural word embedding and through an interactive online platform, ensures transparency and incorporates domain knowledge. The approach demonstrates how machine learning techniques  may be used to enhance the curation process for humanities scholars. Evaluation of the system demonstrated  that domain-specific conceptual modelling with neural word embedding is effective in uncovering texts that capture given concepts.  Texts retrieved using the \emph{Curatr} text mining workflow were particularly useful in the context of exploratory humanities research. The system uncovered works that had not previously attracted sustained scholarly attention, and which presented opportunities to uncover new insights in relation to the given research topic. In future work the addition of new sources to the platform would further enrich this process. Through the exploratory text mining workflow supported by \emph{Curatr}, humanities scholars are enabled to challenge prevailing methods of canonisation of historical fiction.

\vskip 1.3em
\noindent \textbf{Acknowledgements.} This research was partially supported by Science Foundation Ireland (SFI) under Grant Number SFI/12/RC/2289.

\bibliographystyle{splncs04}
\bibliography{sample-base}

\end{document}